\pdfoutput=1

\documentclass[11pt]{article}

\usepackage[preprint]{acl}

\usepackage{times}
\usepackage{latexsym}

\usepackage{graphicx}

\usepackage[T1]{fontenc}

\usepackage[utf8]{inputenc}

\usepackage{microtype}

\usepackage{inconsolata}

\def\eg{\emph{e.g.}}

\def\ie{\emph{i.e.}}

\usepackage{algorithm}
\usepackage{algorithmic}
\usepackage{amsmath}
\usepackage{amssymb}
\usepackage{booktabs}

\title{StyleDubber: Towards Multi-Scale Style Learning for Movie Dubbing}

\author{
Gaoxiang Cong$^{1}$~~~Yuankai Qi$^{2}$\footnotemark[1]~~~Liang Li$^{1}$\footnotemark[1]~~~Amin Beheshti$^{2}$~~~Zhedong Zhang$^{3}$\\~~~\textbf{Anton van den Hengel}$^{4}$
~~~\textbf{Ming-Hsuan Yang}$^{5}$~~~\textbf{Chenggang Yan}$^{3}$~~~\textbf{Qingming Huang}$^{1}$\\
	$^1$Institute of Computing Technology, CAS~~~$^2$Macquarie University~~~\\$^3$Hangzhou Dianzi University~~~$^4$University of Adelaide~~~$^5$University of California~~~\\
 \texttt{\small{conggaoxiang@foxmail.com, yuankai.qi@mq.edu.au, liang.li@ict.ac.cn}}
 }

\begin{document}
\maketitle

\footnotetext[1]{Corresponding authors}

\begin{abstract} 
{
Given a script, the challenge in Movie Dubbing (Visual Voice Cloning, V2C)  is to generate speech that aligns well with the video in both time and emotion, based on the tone of a reference audio track. 
Existing state-of-the-art V2C  models break the phonemes in the script according to the divisions between video frames, which solves the temporal alignment problem but  
leads to incomplete phoneme pronunciation and poor identity stability.  
To address this problem, we propose StyleDubber,
which switches dubbing learning from the frame level to phoneme level. 
It contains three main components:
(1) A multimodal style adaptor operating at the phoneme level to learn pronunciation style
from the reference audio, and generate intermediate representations informed by the facial emotion presented in the video;
(2) An utterance-level style learning module, which guides both the  mel-spectrogram decoding and the refining processes from the intermediate embeddings to improve the overall style expression;
And (3) a phoneme-guided lip aligner to maintain lip sync. 
Extensive experiments on two of the primary benchmarks, V2C and Grid, demonstrate the favorable performance of the proposed method as compared to the current state-of-the-art. 
The code will be made available at {\href{https://github.com/GalaxyCong/StyleDubber}{https://github.com/GalaxyCong/StyleDubber}. 
}}

\end{abstract}

\section{Introduction}

\begin{figure}[t]
    \centering
    \includegraphics[width=1.0\linewidth]{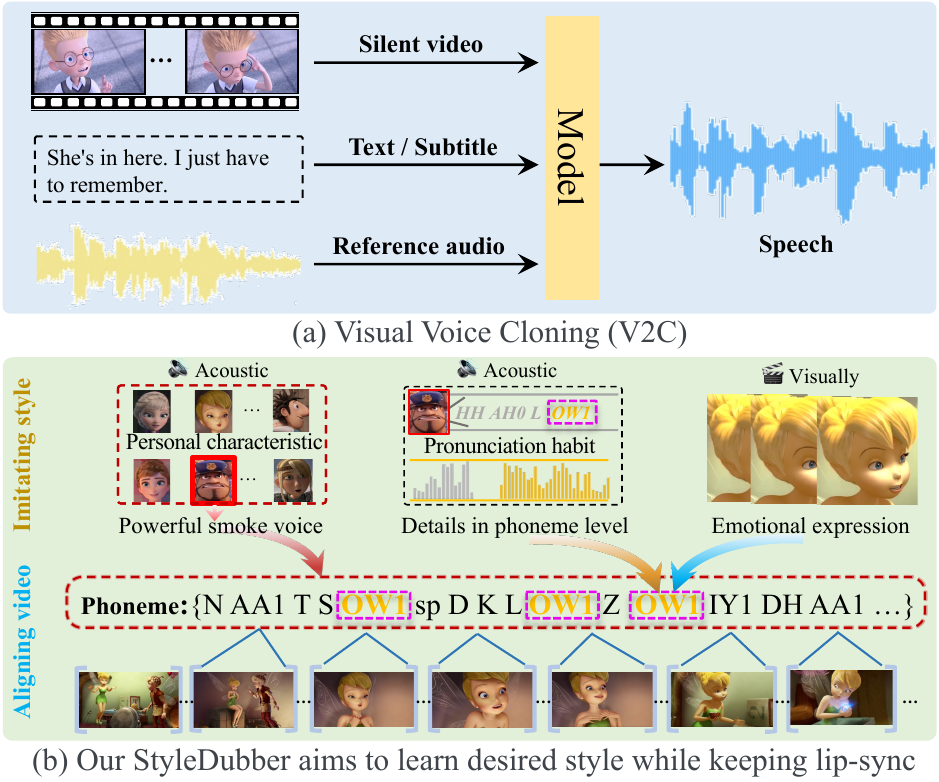}
    \caption{(a) Illustration of the V2C task.
    (b) Our StyleDubber learns  speech styles on two levels: phoneme-level focuses on pronunciation details, 
    while utterance-level emphasizes the overall consistency like timbre. 
    }
    \label{fig1}
\end{figure}

Movie Dubbing~\cite{chen2022v2c}, also known as Visual Voice Cloning (V2C), aims to convert a script into speech with the voice characteristics specified by the reference audio, while maintaining lip-sync with a video clip, and reflecting the character’s emotions depicted therein (see  Figure~\ref{fig1} (a)). 
V2C is more challenging than conventional text-to-speech (TTS)~\cite{ shen2018natural, ren2020fastspeech}, and has obvious applications in the film industry and audio AIGC, including broadening the audience for existing video.

{
Existing methods broadly fall into two groups. 
One group of methods focus primarily on achieving audio-visual sync. 
For example, a duration aligner is introduced in \cite{hu2021neural,cong2023learning} to explicitly control the speed and pause-duration of speaker content by mapping textual phonemes to video frames.  
Then, an upsampling process is used to expand the video frame sequence to the length of mel-spectrogram frame sequence  by multiplying by a  fixed coefficient. 
However, the frame level alignment makes it hard to learn complete phoneme pronunciations, and often leads to seemingly mumbled pronunciations.  
}
The other family of methods focuses on maintaining identity consistency between the generated speech and the reference audio. 
To enable the model to handle a multi-speaker environment,  a speaker encoder is used to extract identity embeddings through averaging and normalizing per speaker embeddings~\cite{hassid2022more}. 
In contrast,~\cite{lee2023imaginary} and~\cite{hu2021neural} try to learn desired speaker voices based on facial appearances. 
Although humans' faces can reflect some vocal attributes (\eg, age and identity) to some extent, they rarely encode speech styles, such as pronunciation habits or accents.

According to~\cite{YixuanZhou,li2021towardsinterspeech}, 
human speech can be perceived as a compound of multi-acoustic factors: 
(1) unique characteristics, such as timbre, which can be reflected on utterance level (see left panel of Figure~\ref{fig1} (b));  
(2) 
pronunciation habits, such as the rhythm and regional accent, which are usually reflected at the phoneme level (see pink rectangles in Figure~\ref{fig1} (b)). 
We also note that one's voice can be affected by emotions. 
For example, the voice can be significantly different when one gets  angry. 
Based on these observations,
we propose to learn phoneme level representations from the speaker's pronunciation habits reflected in the reference audio, and  take both facial expressions and overall timbre characteristics at the utterance level of the reference audio into consideration when generating speech.

In light of the above, we propose 
StyleDubber, 
which learns a desired style 
at the phoneme and utterance levels instead of the conventional video frame level. 
Specifically, a multimodal phoneme adaptor (MPA) is proposed to capture the pronunciation styles at the phoneme level. 
By leveraging the cross-attention relevance between textual phonemes of the script and the reference audio as well as visual emotions, MPA learns the reference style and then generates 
intermediate speech representations with consideration of the required emotion. 
Our model also introduce an utterance-level style learning (USL) module to strengthen personal characteristics during both the mel-spectrogram decoding and refining processes from the above intermediate representations. 
For the temporal alignment between the resulting speech and the video, we propose a Phoneme-guided Lip Aligner (PLA) to synchronize lip-motion and phoneme embeddings. 
At last, HiFiGAN~\cite{Kong2020HiFi} is used as a vocoder to convert the predicted mel-spectrogram to the time-domain waves of dubbing.

The main contributions are summarized as:
\begin{itemize}
    \item 
    We propose StyleDubber, a style-adaptative dubbing model, which imitates a desired personal style in phoneme and utterance levels.  
    It enhances speech generation in terms of speech clarity and its temporal alignment with  video.
    \item At the phoneme level, we design a multimodal style adaptor, which learns styled pronunciation of textual phonemes and considers facial expressions when generating intermediate speech representations. 
    At the utterance level, our model learns to impose timbre into resulting mel-spectrograms. 
    \item Extensive experimental results show that our model performs favorably compared to current state-of-the-art methods.
\end{itemize}

\begin{figure*}[!htbp]
    \centering
    \includegraphics[width=1.0\linewidth]{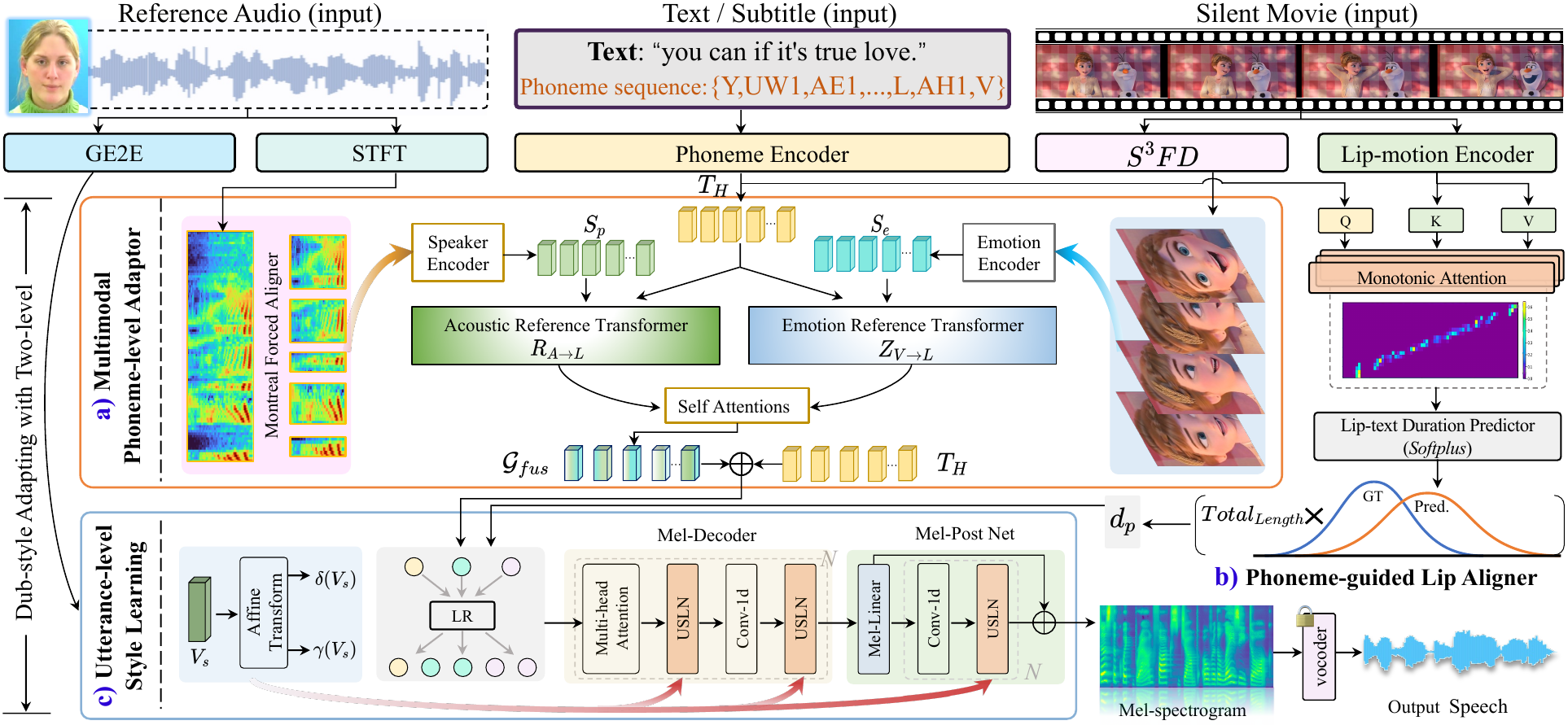}
    \caption{
    The main architecture of the proposed StyleDubber. It consists of a) Multimodal Phoneme-level Adaptor (MPA) (Sec.~\ref{sec:mpa}), b) Phoneme-guided Lip Aligner (PLA) (Sec.~\ref{sec:pla}), and c) Utterance-level Style Learning (USL) (Sec.~\ref{sec:usl}). Note that $\oplus$ is intended to denote vector addition.
    }
    \vspace{-8pt}
    \label{fig3}
\end{figure*}

\section{Related Work}

\textbf{Text to Speech} is a longstanding problem, but recent models represent a dramatic improvement~\cite{RuiLiuACM, XuTanPAMI, EdressonYourTTS, wang2023neural, FastDiffRevisiting, Prosody-TTSDiffusion, ju2024naturalspeech}. 
FastSpeech2~\cite{ren2020fastspeech}, for example, alleviates the one-to-many text-to-speech mapping problem by explicitly modeling variation information.  
\citet{Dongchan2021StyleSpeech}, in contrast, improves  generalization through episodic meta-learning and generative adversarial networks. 
Recently \citet{le2023voicebox} proposed a non-autoregressive flow-matching model for mono or cross-lingual zero-shot text-to-speech synthesis. 
Despite the impressive speech they generate, these methods cannot be applied to the V2C task as they lack the required emotion modelling and lip sync. 

\vspace{1mm}
\noindent\textbf{Visual Voice Cloning} is proposed to address the problem of film dubbing~\cite{chen2022v2c} and has attracted a lot of attention in cross-modality alignment field~\cite{tu20222,tu2023relation, tu2024smart, LiangLong, XuejingEntity, WangHaoPAMI, xiao2023r, xiao2022few}. 
Then, \citet{cong2023learning} proposed a hierarchical prosody dubbing model by associating with lip, face, and scene and focus on frame-level prosody learning~\cite{hu2021neural}. 
To handle multi-speaker scenes, \citet{hassid2022more} matches identities by normalizing each speaker to the unit norm, and \citet{lu2022visualtts} adopts a lookup table to match the d-vector. 
Recently, Face-TTS~\cite{lee2023imaginary} used biometric information extracted directly from the face image as style to improve identity modelling using a score-based diffusion model. 
Unlike the above methods, StyleDubber address the challenge of insufficient identity information by introducing adaptive utterance-level embedding and detailed pronunciation variations based on the reference audio and video. 

\vspace{1mm}
\noindent\textbf{Human Pronunciation Modeling} aims to learn individual pronunciation variations, which is crucial to generate comprehensible, natural, and acceptable speech~\cite{miller1998pronunciation}. 
Compared with fixed speaker representations, phoneme-dependent methods~\cite{li2021towardsinterspeech, RuiboPhoneme} can better control speech and describe more pronunciation features, as phonemes are the basic sound units in a language~\cite{lubis2023basic}. 
Recently, \citet{YixuanZhou} analysed the correlation between local pronunciation content and speaker embeddings at the quasi-phoneme level by reference attention. 
Here, in contrast, we propose a multimodal style adaptor to capture the fine-grained pronunciation variation, which not only imitates the reference style acoustically, but also conveys emotional expression by reference transformer.

\section{Proposed Method}

\subsection{Overview}
Our StyleDubber aims to generate a desired dubbing speech $\hat{Y}$, given a reference audio $R_a$, a phoneme sequence $T_p$ converted from the given script, and a video frame sequence $V_l$: 
\begin{equation}
    \hat{{Y}} = \mathrm{StyleDubber}(R_a, T_p, V_l). 
\end{equation}
The main architecture of the model is shown in Figure~\ref{fig3}. 
Unlike existing prosody dubbing methods, 
our model learns speech style from phoneme level and utterance level,
inspired by human tonal phonetics. 
First, the textual phoneme sequence is converted from raw text. 
A phoneme encoder is then used to extract phoneme embeddings. 
These embeddings are fed into our Multimodal Phoneme-level Adaptor (MPA), which learns to capture and apply phoneme-level pronunciation styles to generate intermediate speech representations, meanwhile taking facial expressions into consideration. 
Next, our Phoneme-guided Lip Aligner (PLA)  predicts the duration for each phoneme by associating lip motion sequence. 
The duration and intermediate dubbing representation are fed to our Utterance-level Style Learning (USL) module, which learns overall style at the utterance level and applys it during mel-spectrograms decoding and refining processes.
We detail each module below.

\subsection{Multimodal Phoneme-level Adaptor}\label{sec:mpa} 
Our Multimodal Phoneme-level Adaptor (MPA) contains three steps: (1) learn acoustic style from reference audio; (2) perceive visual emotion from silent movies; 
(3) generate intermediate speech representations for textual phonemes of the input script, with reference to the captured acoustic styles and emotions in the last two steps.

\smallskip \noindent\textbf{Learn acoustic style.} 
We extract reference mel-spectrogram $R_{mel}$ from reference audio $R_a$ by Short-time Fourier transform (STFT), and the montreal forced aligner~\cite{mcauliffe2017montreal} is used to clip phoneme. 
Then, 
we capture the style feature $S_p$ 
via a mel-style encoder $\mathrm{E_{down}^{spk}}(\cdot)$: 
\begin{equation}
S_{p} = \mathrm{E_{down}^{spk}}({R}_{mel}), 
\end{equation} 
where $\mathrm{E_{down}^{spk}}(\cdot)$ comprises a mel-style encoder~\cite{Dongchan2021StyleSpeech} and four 1D convolutional downsample layers~\cite{YixuanZhou}. 
On the other hand,
embeddings of textual phoneme sequence $T_H\in\mathbb{R}^{{{N_p}}\times{D_m}}$ are extracted by a phoneme encoder $T_H = E_{pho}(T_p)$~\cite{cong2023learning}, where $N_p$ denotes the length of phoneme sequence. 
Next, we propose the acoustic reference transformer $R_{A \rightarrow L}$ (Acoustic to Language) to calculate the relevance between a textual phoneme embedding and 
each style feature by crossmodal transformer: 
\begin{equation}
{\small{
    \begin{split}
        R^{[0]}_{A \rightarrow L} = & \; R^{[0]}_L, \\
        \hat{R}^{[i]}_{A \rightarrow L} = & \; \text{CM}^{[i],\text{mul}}_{A \rightarrow L} (\text{LN}(R^{[i-1]}_{A \rightarrow L}), \text{LN}(R^{[0]}_A))
        + \text{LN}(R^{[i-1]}_{A \rightarrow L}), \\
        R^{[i]}_{A \rightarrow L} = & \; f_{\theta^{[i]}_{A \rightarrow L}} (\text{LN}(\hat{R}^{[i]}_{A \rightarrow L}) + \text{LN}(\hat{R}^{[i]}_{A \rightarrow L}),
    \end{split}
}}
\end{equation} 
where $i = \{1,...,D\}$ denotes the number of feed-forwardly layers, $\small{\text{LN}(\cdot)}$ denotes the layer normalization, and $f_{\theta}$ is a positionwise feed-forward sublayer parametrized by $\theta$. 
$\small{\text{CM}^{[i],\text{mul}}_{A \rightarrow L}}(\cdot)$ is a multihead attention between $S_p$ and $T_H$, as follows: 
\begin{equation}
\small{
    \text{CM}^{[i],\text{mul}}_{A \rightarrow L} =  \mathrm{softmax}(\frac{{T_H} {S_p}^\top}{\sqrt{d_{S_p}}}){S_p},
}
\end{equation} 
where the textual phoneme embedding $T_H$ is used as query and the style feature $S_{p}$ is used as key and value. 
Unlike crossmodal transformer in~\cite{MultimodalHung}, our acoustic reference transformer removes the repeatedly reinforcing and MFCCs frame-level operation, and only focuses on interaction between quasi-phoneme scale of reference audio and script phoneme, which is more conducive to human pronunciation habits.

Unlike using cross-entropy loss as style classifier~\cite{YixuanZhou}, we constrain $\mathrm{E_{down}^{spk}}$ via a style consistency loss:
\begin{equation}
\small{
\mathcal{L}_{spk}=\frac{1}{n}\cdot \sum_j^n (1-\mathrm{cos\_sim}(\phi(T_j),A({S_p})_{j})),
}
\end{equation} 
where $\phi(\cdot)$ is a function outputting the embedding by the pre-trained GE2E model~\cite{AlanPapirGE2E}, $A(S_p)$ outputs a style vector via average pooling, and $\mathrm{cos\_sim}(\cdot)$ is the cosine similarity function. 
$T$ represents the ground truth audio, $n$ is batch size.

\smallskip \noindent\textbf{Perceive visual emotion.} 
We first use the $S^3FD$ model~\cite{zhang2017s3fd} to detect facial region from each frame of video, and then an emotion face-alignment network (EmoFAN)~\cite{toisoul2021estimation} is used to extract emotion features $F_p \in\mathbb{R}^{{{N_v}}\times{D_m}}$ from face regions. 
The $N_v$ represents the number of video frames. 
Similar to style extraction, 
emotional feature $S_{e}$ is obtained by a downsampling equipped encoder: 
\begin{equation}
S_{e} = \mathrm{E_{down}^{emo}}(F_{p}),
\end{equation} 
where 
$S_{e}\in\mathbb{R}^{{{N_{dv}}}\times{D_m}}$ 
and 
$N_{dv}$ is length after down-sample. 
The difference from $\mathrm{E_{down}^{spk}}(\cdot)$ is that 
$\mathrm{E_{down}^{emo}}(\cdot)$  has two  1D convolutional downsample layers. 
Next, an emotion reference transformer $Z_{V \rightarrow L}$ (Visual to Language) is proposed to analyze the correlations between the 
emotional feature and textual phoneme. 
The $Z_{V \rightarrow L}$ has same architecture with $R_{A \rightarrow L}$. The $\small{\text{CM}^{[i],\text{mul}}_{V \rightarrow L} (\cdot)}$ is multihead attention to calculate correlation between $S_e$ and $T_H$: 
\begin{equation}
\small{
    \text{CM}^{[i],\text{mul}}_{V \rightarrow L} =  \mathrm{softmax}(\frac{{T_H} {S_e}^\top}{\sqrt{d_{S_e}}}){S_e},
}
\end{equation} 
where key and value are 
emotional features $S_e$ to assist script phoneme in selecting related visual emotion expression. 
Finally, we regard the output $Z^{D}_{V \rightarrow L}$ and $R^{D}_{A \rightarrow L}$ of the last layers of emotion reference transformer and acoustic reference transformer as context visual emotion and acoustic style, respectively. 
The cross-entropy emotional classification loss $\mathcal{L}_{emo}$ is used to constrain $\mathrm{E_{down}^{emo}}(\cdot)$. 

\smallskip \noindent\textbf{Generate intermediate speech representations.} 
We first concatenate the phoneme-level context visual emotion and acoustic style in channel dimension, and then feed it into self-attention blocks $\mathrm{SA}(\cdot)$ to fuse these embeddings: 
\begin{equation}
\small{
\mathcal{G}_{fus}=\mathrm{SA}([Z^{D}_{V \rightarrow L}, R^{D}_{A \rightarrow L}]), 
}
\end{equation} 
where $\mathcal{G}_{fus}\in\mathbb{R}^{{{N_p}}\times{D_m}}$ is the fused multimodal context embedding 
which is with the same length as the textual phonemes. 
Finally, we combine textual phoneme embedding and multimodal context embedding  $\mathcal{O}_{pho} = \mathcal{G}_{fus} \oplus T_H$, which is viewed as the intermediate dubbing representations.

\subsection{Phoneme-guided Lip Aligner}\label{sec:pla} 
The Phoneme-guided Lip Aligner (PLA) consists of two steps: 1) Monotonic attention is used to learn the contextual aligning feature between lip motion and textual phoneme embedding; 2) Lip-text duration predictor aims to output the duration of each phoneme based on the contextual aligning feature. 

\smallskip \noindent\textbf{Monotonic Attention.} 
The lip-movement hidden representation $L_H= E_{lip}(V_l)\in\mathbb{R}^{{{N_v}}\times{D_m}}$  is obtained using the same lip-motion encoder in~\cite{cong2023learning}. 
Then, we encourage PLA to use textual phoneme embedding to capture related lip motion by multi-head attention with monotonic constraint:  
\begin{equation}
\small{
    C_{lip} 
    =  \mathrm{softmax}(\frac{{T_H} {L_H}^\top}{\sqrt{d_{L_H}}}){L_H},
}
\end{equation}
where the textual phoneme embedding  $T_H$ serves as query, and the lip motion embedding $L_H$ serves as key and value.  
$C_{lip}\in\mathbb{R}^{{{N_p}}\times{D_m}}$
captures the dependency between lip and textual phoneme. 
Monotonic Alignment Loss (MAL)~\cite{MingjianInterspeech} is used to ensure proper alignment over the time: 
\begin{equation}
\small{
\mathcal{L}_{m} = \mathrm{log}(-\frac{\sum_{l=kp-\beta}^{kp+\beta} \sum_{p=0}^{P-1} M_{p,l}}{\sum_{l=0}^{L-1} \sum_{p=0}^{P-1} M_{p,l}}),
}
\end{equation}
where $\beta$ is a hyper-parameter to control bandwidth, $k$ is the slop for length of phoneme $P$ and corresponding of lip length $L$, and $M_{p,l}$ is the masked attention weight matrix with $p$-th row and $l$-th column. 
To this end, $\mathcal{L}_{mon}$ aims to constrain attention weights to diagonal area to satisfy monotonicity. 

\begin{table*}[!t]
  \centering
  \resizebox{1.0\linewidth}{!}
  {
    \begin{tabular}{c|c|ccccc|ccccc}
    \hline
    Dataset & & \multicolumn{5}{c|}{V2C-Animation} & \multicolumn{5}{c}{GRID} \\ 
    \toprule
    Methods & Visual
    & SPK-SIM (\%) $\uparrow$ 
    & WER (\%) $\downarrow$
    & EMO-ACC (\%) $\uparrow$   
    & MCD-DTW $\downarrow$ 
    & MCD-DTW-SL $\downarrow$
    & SPK-SIM (\%) $\uparrow$ 
    & WER (\%) $\downarrow$
    & EMO-ACC (\%) $\uparrow$   
    & MCD-DTW $\downarrow$ 
    & MCD-DTW-SL $\downarrow$ \\ 
    \midrule
    GT & - & 100.00 & 22.55  & 99.96 & 0.0 & 0.0 & 100.00 & 22.41 & - & 0.0 & 0.0\\
    GT Mel + Vocoder & - & 96.96 & 24.58 & 97.09  & 3.77 & 3.80 & 97.57 & 21.41 & -& 4.10 & 4.15  \\
    \midrule
    Fastspeech2~\cite{ren2020fastspeech} & \text{\sffamily X} & 24.87   & 34.48  & 42.21   &  11.20 & 14.48 &  47.41 & 19.05 &- & 7.67 & 8.43  \\
    StyleSpeech~\cite{Dongchan2021StyleSpeech} & \text{\sffamily X}& 54.99 & 106.73 & 44.12 &  11.50 & 15.10  & 91.06 & 24.83 &-  & 5.87 & 5.98 \\
    Zero-shot TTS~\cite{YixuanZhou} & \text{\sffamily X}& 48.98 & 68.81 & 42.75 & 9.98 & 12.51 & 86.54 & 19.13 & -& 5.71 & 5.99\\
    StyleSpeech*~\cite{Dongchan2021StyleSpeech} & \checkmark & 42.53 & 108.00 & 42.53   & 11.62 & 14.23 &  90.04 &22.62 & - & 5.74 & 5.88  \\
    Fastspeech2*~\cite{ren2020fastspeech} & \checkmark &  25.47   & 33.53 & 42.39  &  11.35 & 14.73 & 59.58  & 19.61 & - & 7.24 & 7.95 \\
    Zero-shot TTS*~\cite{YixuanZhou} & \checkmark & 48.93  & 68.05 & 43.97 & 10.03  &  12.01 & 85.93 & 20.05 & -  & 5.75 &  6.40 \\
    V2C-Net~\cite{chen2022v2c} & \checkmark & 40.61   & 73.08 & 43.08 & 14.12 & 18.49 & 80.98  & 47.82 & - & 6.79 & 7.23 \\
    HPMDubbing~\cite{cong2023learning} & \checkmark & 53.76  & 164.16 & \textbf{46.61} & 11.12  & 11.22 & 85.11 & 45.11 & - & 6.49 & 6.78 \\
    Face-TTS~\cite{lee2023imaginary} & \checkmark & 52.81  & 201.13  & 44.04 &  13.44 & 26.94 & 82.97 & 44.37 & -& 7.44 &  8.16   \\
    \midrule
     Ours & \checkmark &  \textbf{82.26}    &  \textbf{31.49}   &  {45.62}  &  \textbf{9.37}  &  \textbf{9.46}  &  \textbf{93.79}  &  \textbf{18.88}  &   - & \textbf{5.61} & \textbf{5.69} \\
    \bottomrule
    \end{tabular}
    }
    \vspace{-5pt}
  \caption{
  {
  Results under the Dub 1.0 setting~\cite{chen2022v2c}, which uses ground-truth audio as reference audio. The method with ``*'' refers to a variant taking video embedding as an additional input as in~\cite{chen2022v2c}. 
The metric EMO-ACC is not applicable to GRID as it does not have emotional labels.
  }
  }
  \vspace{-5pt}
  \label{tab1}
\end{table*}

\begin{table*}[!t]
  \centering
  \resizebox{1.0\linewidth}{!}
  {
    \begin{tabular}{c|c|cccccccc}
    \toprule
    Methods & Visual
    & SPK-SIM (\%) $\uparrow$ 
    & WER (\%) $\downarrow$
    & EMO-ACC (\%) $\uparrow$   
    & MCD-DTW $\downarrow$ 
    & MCD-DTW-SL $\downarrow$
    & MOS-similarity $\uparrow$ 
    & MOS-naturalness $\uparrow$   \\
    \midrule
    GT & - & 100.00 & 22.76  & 99.96 & 0.0 & 0.0 & 4.69 $\pm$ 0.12 & 4.76 $\pm$ 0.09 \\
    GT Mel + Vocoder & - & 96.93 & 24.83  &  96.95 & 3.77 & 3.80 & 4.65 $\pm$ 0.07 & 4.63$\pm$  0.09 \\
    \midrule
    Fastspeech2~\cite{ren2020fastspeech} & \text{\sffamily X} & 24.17   & 35.08  & 42.21   &  11.20 & 14.48 & 2.13$\pm$ 0.09  & 3.75$\pm$ 0.12 \\
    StyleSpeech~\cite{Dongchan2021StyleSpeech} & \text{\sffamily X}& 75.66 & 76.58 & 41.55 &  11.56 & 15.10  & 3.35 $\pm$ 0.07 & 3.24$\pm$ 0.08 \\
    Zero-shot TTS~\cite{YixuanZhou} & \text{\sffamily X}& 47.79 & 58.82 & 39.11 & 10.68 & 13.52 & 3.58 $\pm$0.11 & 3.72$\pm$ 0.15 \\
    StyleSpeech*~\cite{Dongchan2021StyleSpeech} & \checkmark & 75.67 & 82.48 & 42.57   & 11.58 & 15.23 & 3.46 $\pm$ 0.16 & 3.83$\pm$ 0.15 \\
    Fastspeech2*~\cite{ren2020fastspeech} & \checkmark &  25.47   & 34.08 & 42.39  &  11.35 & 14.73 & 2.46 $\pm$0.06  & 3.77 $\pm$0.08  \\
    Zero-shot TTS*~\cite{YixuanZhou} & \checkmark &  47.55 & 58.81  & 39.30  & 10.76 & 13.66 & 3.68$\pm$ 0.14 & 3.69 $\pm$0.09 \\
    V2C-Net~\cite{chen2022v2c} & \checkmark &   34.07  &  61.61 & 41.01 & 14.58 & 18.73 & 3.04 $\pm$0.15 & 2.78 $\pm$0.06 \\
    HPMDubbing~\cite{cong2023learning} & \checkmark & 31.42  & 171.03 & \textbf{43.97}  & 11.88 & 11.98 & 3.19 $\pm$0.10 & 3.06 $\pm$0.22\\
    Face-TTS~\cite{lee2023imaginary} & \checkmark & 51.98  & 200.18 & 43.56 &  13.78 & 28.03 & 3.13$\pm$ 0.12 &  3.09 $\pm$0.06  \\
    \midrule
     Ours & \checkmark &  \textbf{81.27}    &  \textbf{31.70}   &  {41.35}  &  \textbf{10.59}  &  \textbf{10.68}  &  \textbf{3.92 $\pm$0.11}  &  \textbf{3.86 $\pm$0.09} \\
    \bottomrule
    \end{tabular}
    }
    \vspace{-5pt}
  \caption{
  {
  V2C-Animation results under Dub 2.0 setting, which uses non-ground truth audio of the desired character as reference audio.
  }
  }
  \vspace{-10pt}
  \label{tab2_V2C}%
\end{table*}%

\smallskip \noindent\textbf{Duration predictor}.  
Since the total dubbing times $Total_{Length}$ can known by multiplying time coefficient with video frames $N_v$ in advance~\cite{hu2021neural}, we transform the alignment problem into inferring the relative time of a phoneme over its total duration. 
We first use the duration predictor~\cite{ren2020fastspeech} to learn the duration from lip-phoneme context $C_{lip}$ and re-scale it into relative duration by using $Total_{Length}$ divide predicted sum: 
\begin{equation}
\small{
    {d}_p = Total_{Length} \cdot \frac{\mathrm{E_{Softplus}}(C_{lip}^k)}{ \sum_{k=0}^{N_p-1} \mathrm{E_{Softplus}}(C_{lip})},
}
\end{equation}
where $d_p \in\mathbb{R}^{{{N_p}}\times{1}}$ represents the relative duration for each phoneme unit. $\mathrm{E_{Softplus}(\cdot)}$ represents the duration predictor, which consist of 2-layer 1D convolutional with softplus activate function~\cite{SongZhengchenDian}. 
In this case, we obtain how many mel-frames correspond to lip-phoneme context $C_{lip}$ to ensure the boundary of phoneme unit will not be broken, while syncing with the whole video.

\smallskip \noindent\textbf{Loss function}. 
The duration loss is optimized with MSE loss, following~\cite{ren2020fastspeech}: 
\begin{equation}
\mathcal{L}_{d} = \mathrm{MSE}(d_p, \mathrm{log}(g_d)),
\end{equation}
where $\mathrm{log}(g_d)$ represents the ground-truth duration in the log domain. 

\vspace{1mm}
\subsection{Utterance-level Style Learning}\label{sec:usl} 
We also consider the utterance-level information of reference audio to enhance global style characteristics. 
Specifically, we use the GE2E model~\cite{AlanPapirGE2E} to extract the timbre vector $V_{s}$ as utterance-level condition, which aggregates global style information to guide the decoding and refinement of mel-spectrograms from intermediate speech representations by affine transform.

\smallskip \noindent\textbf{Mel-Decoder}. 
We use transformer-based mel-decoder~\cite{cong2023learning} to decode intermediate speech representations $\mathcal{O}_{pho}$ into a spectrogram hidden sequence: 
\begin{equation}
\hat{{R}} = \mathrm{Decoder_{USLN}}(\mathrm{LR}(\mathcal{O}_{pho}, d_p), V_{s}),%
\end{equation}
where $\mathrm{LR}(\cdot)$ is the length regulator~\cite{ren2020fastspeech} to expand $\mathcal{O}_{pho}$ to mel-length based on predicted duration $d_p$.  
$\hat{{R}}\in\mathbb{R}^{{{N_{lr}}}\times{256}}$ denotes a spectrogram hidden sequence, $N_{lr}$ is the predicted total mel-legnth. 
During decoding, we replace  the original layer norm~\cite{ba2016layer} in each Feed-Forward Transformer (FFT) block with our utterance-level style learning normalization (USLN): 
\begin{equation}
\mathrm{USLN}(h, V_{s}) = \gamma(V_{s}) \cdot h_n + \delta(V_{s}),
\end{equation}
where $h_n = (h - \mu)/ {\sigma}$ is normalized features by the mean $\mu$ and variance $\sigma$ of input feature $h$. 
The $\gamma(V_{s})$ and $\delta(V_{s})$ represent the learnable gain and bias of the overall style vector by affine transform, respectively, which can adaptively perform scaling and shifting to improve style expression.

\smallskip \noindent\textbf{Refine mel-spectrogram}. We introduce the aforementioned USLN to MelPostNet~\cite{JonathanShenICASSP} to inject the style information from timbre vector $V_{s}$  
during refining the final mel-spectrograms stage: 
\begin{equation}
\small{
\hat{{M}} = \mathrm{POST_{USLN}}(\hat{{R}}, V_{s}),
}
\end{equation}
where $\hat{{M}}\in\mathbb{R}^{{{N_{lr}}}\times{80}}$ denotes the predicted mel-spectrograms with 80 channels.

\subsection{Training} 
Our model is trained in an end-to-end fashion via optimizing the sum of all losses. 
The total loss $\mathcal{L}$ can be formulated as: 
\begin{equation}
\small
{
\mathcal{L}=\lambda_1 \mathcal{L}_{spk}+\lambda_2 {L}_{emo} + \lambda_3 \mathcal{L}_{r}+\lambda_4 \mathcal{L}_{m}+\lambda_5 \mathcal{L}_{d},
}
\label{eq:loss_all}
\end{equation}
where $\mathcal{L}_{r}$ is the reconstruction loss to calculate $\mathrm{L1}$ differences between the predicted and ground-truth mel-spectrograms. 

Finally, the generated mel-spectrograms $\hat{{M}}$ are converted to time-domain wave $\hat{{Y}}$ via the widely used vocoder HiFiGAN.

\section{Experiments} 
We evaluate our method on two primary V2C datasets, V2C-Animation and GRID. Below, we first provide implementation details. Then, we briefly introduce the datasets and evaluation metrics, followed by quantitative and qualitative results. Ablation studies are also conducted to thoroughly evaluate our model.

\subsection{Implementation Details}
The video frames are sampled at 25 FPS and all audios are resampled to 22.05kHz. 
The ground-truth of phoneme duration is extracted by montreal forced aligner~\cite{mcauliffe2017montreal}. 
The window length, frame size, and hop length in STFT are 1024, 1024, and 256, respectively. 
The lip region is resized to 96 $\times$ 96 and pretrained on ResNet-18, following~\cite{martinez2020lipreading, ma2020towards}.  
We use 8 heads for multi-head attention in PLA and the hidden size is 512. 
The duration predictor consists of 2-layer 1D convolution
with kernel size 1. 
The  weights in {Eq.~\ref{eq:loss_all}} are set to $\lambda_1$ = 25.0, $\lambda_2$ = 0.1, $\lambda_3$ = 5.0, $\lambda_4$ = 2.0, $\lambda_5$ = 5.0. 
For downsampling encoder in $\mathrm{E_{down}^{spk}}$, we use 4 convolutions containing [128, 256, 512, 512] filters with shape 3 $\times$ 1 respectively, each followed by an average pooling layer with kernel size 2. 
In $\mathrm{E_{down}^{emo}}$, 2 convolutions are used to download to quasi phoneme-level, containing [128, 256] filters with shape 3 $\times$ 1. 
In $R_{A \rightarrow L}$ and $Z_{V \rightarrow L}$, the dimensionality of all reference attention hidden size is set to 128 implemented by a 1D temporal convolutional layer.  
We set the batch size to 32 and 64 on V2C-Animation and Grid dataset, respectively. 
For training, we use Adam~\cite{kingma2014adam} optimizer with $\beta_1$ = 0.9, $\beta_2$ = 0.98, $\epsilon$=$10^{-9}$ to optimize our model. 
The learning rate is set to 0.00625. 
Both training and inference are implemented with PyTorch on a GeForce RTX 4090 GPU. 

\subsection{Dataset}
{
\noindent\textbf{V2C-Animation dataset}~\cite{chen2022v2c} is currently the only publicly available movie dubbing dataset for multi-speaker. 
Specifically, it contains 153 diverse characters extracted from 26 Disney cartoon movies, specified with speaker identity and emotion annotations. 
The whole dataset has 10,217 video clips, and the audio samples are sampled at 22,050Hz with 16 bits. 
In practice,~\cite{chen2022v2c} removes video clips less than 1s. 
In this work, all experiments are conducted on the V2C denoise version. 
We will publish this version. 
}

\noindent\textbf{GRID dataset}~\cite{cooke2006audio} is a basic benchmark for multi-speaker dubbing.  
The whole dataset has 33 speakers, each with 1000 short English samples. 
All participants are recorded in a noise-free studio with a unified screen background. 
The train set consists of 32,670 samples, 900 sentences from each speaker. 
In the test set, there are 100 samples of each speaker.

\subsection{Evaluation Metrics} 
\noindent\textbf{Objective metrics}. 
{
To measure whether the generated speech carries the desired speaker identity and emotion, speaker identity similarity (SPK-SIM) is calculated by SECS~\cite{SC_Glow}, and emotion accuracy (EMO-ACC) is employed by pre-trained speech emotion recognition model~\cite{emo_acc_Jiaxin}. 
Besides, we adopt the Mel Cepstral Distortion Dynamic Time Warping (MCD-DTW) to measure the difference between generated speech and real speech. 
We also adopt the metric MCD-DTW-SL, which is MCD-DTW weighted by duration consistency~\cite{chen2022v2c}. 
}
The Word Error Rate (WER)~\cite{AndrewWER} is used to measure pronunciation accuracy by the publicly available whisper~\cite{whisper} as the ASR model. 
Besides, we use the ASV model (WavLM-TDNN~\cite{SanyuanWavLM}) to comprehensively evaluate identity similarity (see Appendix~\ref{spksim}), following NaturalSpeech 3~\cite{ju2024naturalspeech}.

\vspace{1mm}
\noindent\textbf{Subjective metrics}. 
{
We also provide subjective evaluation results via conducting a human study using a 5-scale mean opinion score (MOS) in two aspects: naturalness and similarity. 
Following the settings in \cite{chen2022v2c}, all participants are asked to assess the sound quality of 25 randomly selected audio samples from each test set. 
}

\subsection{Performance Evaluations} 
\begin{table}[!tbp]
  \centering
  \resizebox{1.0\linewidth}{!}
  {
    \begin{tabular}{llcccccc}
    \toprule
    Setting & Explanation & Num. \\
    \midrule
    Dub 1.0 (Original setting~\cite{chen2022v2c}) &  Ground-truth speaker in test set  & 2,779 \\
    Dub 2.0 (Reference speaker setting)    & Same-speaker from other movie clips & 2,626  \\
    Dub 3.0 (Unseen speaker setting)  &  Unseen speaker & 4,851 \\
    \bottomrule
    \end{tabular}%
    } 
  \caption{Experimental settings for dub testing in V2C.} 
  \vspace{-10pt}
  \label{tab_1_Explanation}%
\end{table}

We evaluate our method in three experimental settings as shown in Table~\ref{tab_1_Explanation}. 
{
The first setting is the same as in~\cite{chen2022v2c}, which uses target audio as reference audio from test set.  
However, this is impractical in real-world applications. 
Thus, we design two new and more reasonable settings:  
``Dub 2.0'' uses non-ground truth audio of the same speaker as  reference audio;
}
``Dub 3.0'' uses the audio of unseen characters (from another dataset) as reference audio. 
We compare with six recent related baselines to comprehensively analyze. 
Furthermore, we will release the detailed configuration for all experiment settings for the GRID and V2C Animation datasets. 

\vspace{1mm}
\noindent\textbf{Results under Dub 1.0 setting}. 
{
As shown in Table~\ref{tab1}, our method achieves the best performance on almost all metrics on both GRID and V2C-Animation benchmarks.  Our method only performs slightly worse in terms of EMO-ACC  than the SOTA movie dubbing model HPMDubbing~\cite{cong2023learning}.  
Regarding identity accuracy (see SPK-SIM), our method outperforms other models with an absolute margin of }
27.27\% 
{
over the 2nd best method. 
} 
In terms of MCD-DTW and MCD-DTW-SL, our method achieves  6.11\% and 24.38\% improvements, respectively. 
This indicates our method can achieve better speech quality and better duration consistency.

\vspace{1mm}
\noindent\textbf{Results under Dub 2.0 setting}. We report the V2C results in Table~\ref{tab2_V2C}. 
{
Despite Dub 2.0 is much more challenging than 1.0, our method still outperforms other methods on six metrics. 
The SPK-SIM and WER significantly improve 7.4\% and 6.98\% than the fastspeech 2 (TTS-textonly) and meta-stylespeech method, respectively. 
}
Additionally, the proposed method (StyleDubber) achieves the lowest MCD-DTW compared to all baselines, which indicates our method achieves minimal acoustic difference even in challenging setting 2.0.  
Furthermore, the lowest MCD-DTW-SL shows that our method achieves almost the same duration sync as the groud-truth video. 
{
Finally, the human subjective evaluation results (see MOS-N and MOS-S) also show that our StyleDubber can generate speeches that are closer to realistic speech in both naturalness and similarity. 
}

\begin{table}[!tbp]
  \centering
  \resizebox{1.0\linewidth}{!}
  {
    \begin{tabular}{lccccccc}
    \toprule
    Methods & Visual & MOS-S & MOS-N& SPK-SIM & WER  \\
    \midrule
    Fastspeech2~\cite{ren2020fastspeech} & \text{\sffamily X} & 2.91 $\pm$ 0.13 & 3.02 $\pm$ 0.09  & 21.11 & 14.05 \\
    StyleSpeech~\cite{Dongchan2021StyleSpeech} & \text{\sffamily X} &3.17 $\pm$ 0.06 & 3.22 $\pm$ 0.15  &55.81& 77.46\\
    Zero-shot TTS~\cite{YixuanZhou} & \text{\sffamily X}  & 3.53 $\pm$ 0.12 & 3.35 $\pm$ 0.07  & 57.23 & 19.83\\
    \midrule
    Fastspeech2*~\cite{ren2020fastspeech} &  \checkmark  &2.97 $\pm$ 0.12 &3.03 $\pm$ 0.29 &26.79& 18.38\\ 
    StyleSpeech*~\cite{Dongchan2021StyleSpeech} &  \checkmark   &3.31 $\pm$ 0.18  &3.22 $\pm$ 0.10 &58.71& 89.11\\
    Zero-shot TTS*~\cite{YixuanZhou}  &  \checkmark   & 3.62 $\pm$ 0.09 & 3.31 $\pm$ 0.13  & 61.12 & 19.25\\
    V2C-Net~\cite{chen2022v2c}  & \checkmark  & 3.05 $\pm$ 0.07
     & 2.83 $\pm$ 0.09 & 38.43 & 112.71\\
    HPMDubbing~\cite{cong2023learning} &  \checkmark  & 3.11 $\pm$ 0.08
     & 2.92 $\pm$  0.09 & 49.31 & 106.81 \\
    Face-TTS~\cite{lee2023imaginary} & \checkmark   & 3.10 $\pm$ 0.05 & 3.17 $\pm$ 0.15   & 33.80 & 201.98\\
     \midrule 
    Ours  & \checkmark & \textbf{3.94$\pm$ 0.12} & \textbf{3.87$\pm$ 0.14} & \textbf{71.52} & \textbf{13.45} \\
    \bottomrule
    \end{tabular}%
    } 
  \caption{The V2C results under Dub 3.0 setting, which use unseen speaker as refernce audio.} 
  \vspace{-10pt}
  \label{tab_setting3_Explanation}%
\end{table}

\begin{figure*}[t]
    \centering
    \includegraphics[width=1.0\linewidth]{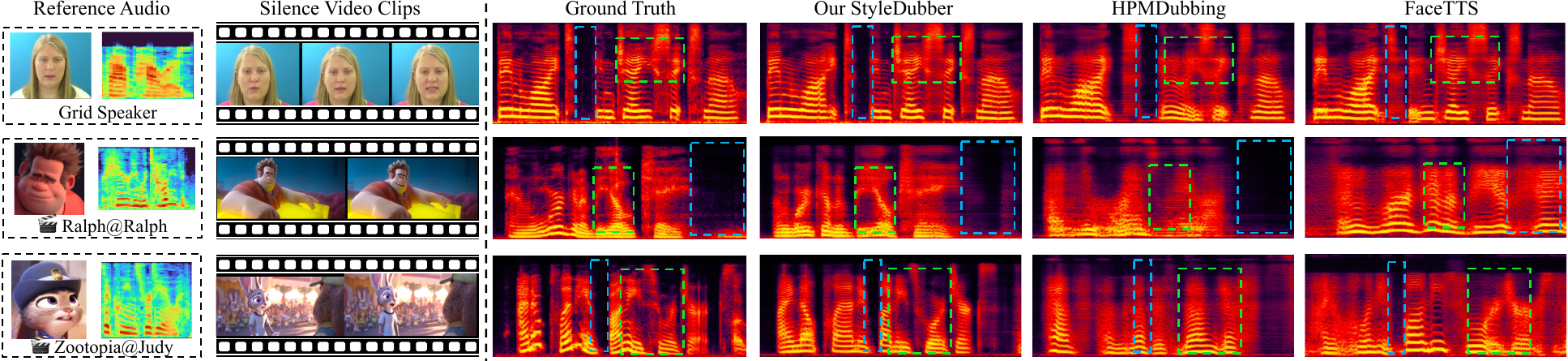}
    \caption{Mel-spectrograms of four synthesized audio samples under the Dub 2.0 setting.   The green and blue rectangles highlight key regions that have significant differences in reconstruction details and duration pause. 
    }
    \vspace{-10pt}
    \label{fig4}
\end{figure*}
\vspace{1mm}

\noindent\textbf{Results under Dub 3.0 setting}. 
Since there is no target audio at this setting, we only compare SPK-SIM and WER, and make subjective evaluations. 
As shown in Table~\ref{tab_setting3_Explanation}, our StyleDubber achieves the best generation quality in all four metrics, largely outperforming the baselines. 
The higher SPK-SIM and MOS-S (mean opinion score of similarity) indicate the better generalization ability of our methods to learn style adation across unseen speakers. 
Besides, our method also maintains good pronunciation 
(see WER). 
Overall, our StyleDubber achieves impressive results in challenging  scenarios.

\subsection{Qualitative Results} 
We visualize the mel-spectrogram of reference audio, ground-truth audio, and synthesized audios by ours and the other two state-of-the-art methods in Figure~\ref{fig4}.  
We highlight the regions in mel-spectrograms where significant differences are observed among these methods in reconstruction details (green boxes), and pause (blue boxes), respectively. 
We find that our method is more similar to the ground-truth mel-spectrogram, which has clearer and distinct horizontal lines in the spectrum to benefit the fine-grained pronunciation expression of speakers (see green box). 
By observing the blue boxes, we find that our method can learn natural pauses to achieve better sync by aligning phonemes and lip motion. 

\begin{table}[!tbp]
  \centering
  \resizebox{1.0\linewidth}{!}
  {
    \begin{tabular}{lccccccc}
    \toprule
    \# & Methods   & WER $\downarrow$ & SPK-SIM $\uparrow$ & MCD-DTW $\downarrow$ & MCD-DTW-SL $\downarrow$ & EMO-ACC $\uparrow$  \\
    \midrule
    1 & w/o MPA  & 39.74  & 77.26 & 9.90  & 9.98 & 41.18  \\
    2 & w/o USL   & 32.81  & 47.07 & 10.23 & 10.43 & 42.77  \\
    3 & w/o PLA & 35.90  & 80.77 & 9.59   & 11.47 & 45.48 \\
    \midrule
    4 & Quasi-phoneme $v.s.$ frame & 33.62  & 80.76 & 9.75    & 9.84 & 43.83 \\
    \midrule
    5 & w/o $R_{A \rightarrow L}$ &  38.21 & 77.79  & 9.81  & 9.90 & 42.61 \\
    6 & w/o $Z_{V \rightarrow L}$  & 33.87 & 79.30  &9.82  & 9.91 & 41.49\\
    \midrule
    7 & w/o U-MelDecoder & 32.26  & 47.31  &  10.31  & 10.41 & 42.80 \\
    8 & w/o U-Post  &  31.73 & 80.79  & 9.52  & 9.61 & 44.33 \\
    \midrule
    9 & Full model  & \textbf{31.49}    &  \textbf{82.26}   &  \textbf{9.37} & \textbf{9.46}  & \textbf{45.62}  \\ 
    \bottomrule
    \end{tabular}%
    } 
  \caption{Ablation study of the proposed method on the V2C benchmark dataset with 1.0 setting, respectively.} 
  \vspace{-10pt}
  \label{tab_ablation}%
\end{table}

\subsection{Ablation Studies}~\label{sec:ag} 

\noindent{To further study the influence of the individual components in StyleDubber, we perform the comprehensive ablation analysis using V2C 1.0 version. } 

\noindent{\textbf{Effectiveness of MPA, USL, and PLA}}. 
{
The results are presented in Row 1-3 of Table~\ref{tab_ablation}. 
It shows that all these three modules contribute significantly to the overall performance, and each module has a different focus. 
}
After removing the MPA, the MCD-DTW and WER severely drop. 
This reflects that the MPA achieves minimal difference in acoustic characteristics from the target speech and better pronunciation by phoneme modeling with other modalities. 
In contrast, the SPK-SIM is most affected by USL, which indicates decoding mel-spectrograms by introducing global style is more beneficial to identity recognition. 
{
Finally, the performance of MCD-DTW-SL drops the most when removing the PLA. 
This can be attributed to the better alignment between video and phoneme sequences. 
}

\vspace{1mm}
\noindent{\textbf{Quasi-phoneme $v.s.$ frame}. 
To prove the impact of regulating the temporal granularity to quasi-phoneme-scale, we remove the downsample operation and retrain the frame-level information as input of $R_{A \rightarrow L}$ and $Z_{V \rightarrow L}$. 
As shown in Row 4 of Table~\ref{tab_ablation}, all metrics have some degree of degradation, which means quasi-phoneme level acoustic and emotion representation is more conducive for script phoneme to capture desired information. 

\vspace{1mm}
\noindent{\textbf{Effectiveness of $R_{A \rightarrow L}$ and $Z_{V \rightarrow L}$}}. 
To study the effect on each reference transformer in MPA, we remove $Z_{V \rightarrow L}$ and $Z_{A \rightarrow L}$, respectively.  
As shown in Row 5-6 of Table~\ref{tab_ablation}, $Z_{V \rightarrow L}$ has a significant effect on improving emotions, while $Z_{A \rightarrow L}$ more focus on local acoustic information to strengthen style and pronunciation. 

\vspace{1mm}
\noindent{\textbf{Effectiveness of U-MelDecoder and U-post}}. 
To prove the effect of each module in USL, we remove the utterance-level style learning on mel-decoder and post-net, respectively. 
In other words, it still keeps an autoregressive manner by transformer-based decoder,  and we just cut off the red arrow in Figure~\ref{fig3} (c). 
As shown in Row 7-8 of Table~\ref{tab_ablation}, when removing the U-post, the performance also drops but is not as large as removing the U-MelDecoder. 
This indicates that U-MelDecoder is critical to the generation of spectrum, while U-post only works on refining spectrum in 80 channels so that the impact is relatively small.

\section{Conclusion}
In this work, we propose StyleDubber for movie dubbing, which imitates the speaker's voice at both phoneme and utterance levels while aligning with a reference video. 
StyleDubber uses a multimodal phoneme-level adaptor to improve pronunciation that captures speech style while considering the visual emotion. 
{
Moreover, a phoneme-guided lip aligner is devised to synchronize vision and speech without destroying the phoneme unit.
The proposed model sets new state-of-the-art on the V2C and GRID benchmarks under three settings.  
}

\section{Limitation} 
We follow the task definition of Visual Voice Cloning (V2C), which focuses on generating audio only. 
Truly solving the larger problem would require changing the video to reflect the updated audio.
In future, we will add this capability to better support tasks like cross-language video translation.

\section{Ethics Statement}  
The existence of V2C methods lowers the barrier to high-quality and expressive visual voice cloning. In the long term this technology might enable broader consumption of factual and fictional video content. This could have employment implications, not least for current film voice actors. 
There is also a risk that V2C might be used to generate fake video depicting people apparently saying things they have never said.  This is achievable already by an impersonator using entry-level video editing software, so the marginal impact of V2C on this problem is small.
The licence for StyleDubber will explicitly prohibit this application, but the efficacy of such bans is limited, not least by the availability of other software that achieves the same purpose. 

\section*{Acknowledgements}
This work was supported in part by National Key R\&D Program of China under Grant (2023YFB4502800), National Natural Science Foundation of China: 62322211, 61931008, 62236008, 62336008, U21B2038, 62225207, Fundamental Research Funds for the Central Universities (E2ET1104), ``Pionee'' and ``Leading Goose'' R\&D Program of Zhejiang Province (2024C01023, 2023C01030). Yuankai Qi, Amin Beheshti, Anton van den Hengel, and Ming-Hsuan Yang are not sup-
ported by the aforementioned fundings.

\bibliography{custom}

\clearpage 
\newpage
\appendix
\noindent {\Large \bf Appendix}\\

We organise the supplementary materials as follows.
\vspace{-5pt}
\begin{itemize}
    \item In Section~\ref{Details_Ch}, we analyze the challenges of the V2C benchmark compared with the traditional TTS benchmark. 
    \vspace{-8pt}
    \item In Section~\ref{Details_baseline}, we introduced related baseline methods.
    \vspace{-8pt}
    \item In Section~\ref{Whisper}, we report the WER result by different Whisper versions on the V2C-Animation dataset. 
    \item In Section~\ref{spksim}, we report the speaker similarity result by the large WavLM-TDNN model on the V2C-Animation dataset and GRID dataset. 
    
    \vspace{-8pt}
\end{itemize}

\section{The challenges in V2C benchmark}\label{Details_Ch}

The V2C benchmark significantly differs from traditional TTS benchmark , and it is more challenging in the following aspects: 
(1) The data scale of V2C dataset is much smaller in terms of either the number of data items or speech length (see Figure~\ref{fig_appendix_Cha} (a)-(b)). 
There are only 9374 video clips in V2C, and most of its audio is shorter than 5s. 
In contrast, FS2 and Stylespeech are trained on LJspeech and LibriTTS with 13,100 and 149,753 samples, most of which are longer than 5s. 
Although LJspeech also looks similar in size to V2C, it is a single-speaker dataset, so V2C allocates fewer samples to each speaker. 
(2) 
V2C has the largest variance of pitch compared to TTS tasks due to exaggerated expressions of cartoon characters (see Figure 1 (c) and more details in Tab. 2 of V2C-Net).
(3)  
The audio of V2C contains background noise or music, like car whistle and alarm clock sound, \emph{et al}. 
Signal-to-noise Ratio (SNR) of V2C is the lowest (Figure~\ref{fig_appendix_Cha} (d)). 
In summary, unlike the large-scale clean TTS datasets, V2C is much more challenging, and the well-known TTS methods suffer performance degradation. 
In this work, all experiments are conducted on the V2C denoise version. 
We will publish this version. 
\begin{figure}[!htbp]
    \centering
    \includegraphics[width=1.0\linewidth]{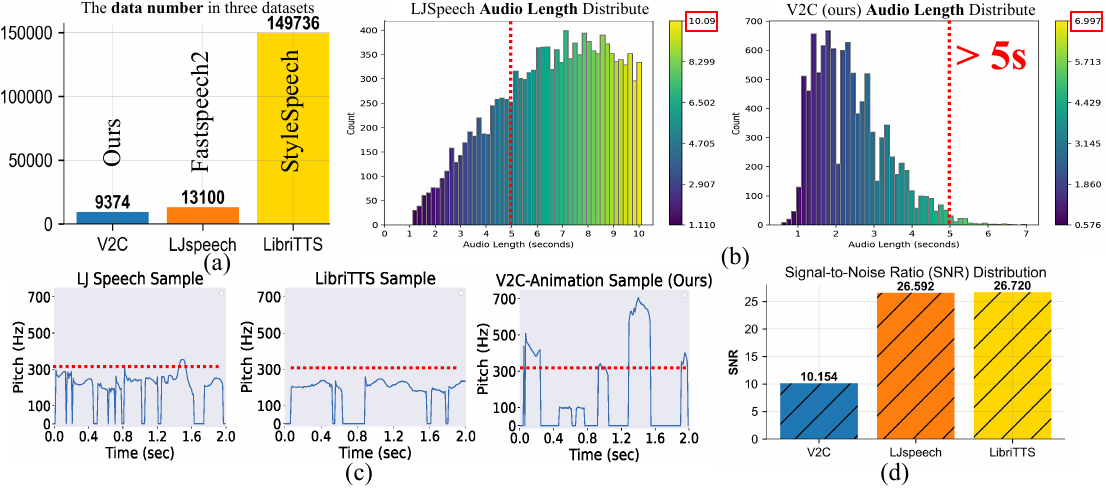}
    \caption{V2C dataset is more challenging than TTS-baseline datasets: (a) fewer samples (only 6567 for training), (b) shorter duration (mostly smaller than 5s), (c) greater variance (pitch), (d) more noise (background sound and music).}
    \vspace{-8pt}
    \label{fig_appendix_Cha}
\end{figure}

\section{Baselines}\label{Details_baseline}
\label{sec:appendix}
We compare our method against six closely related methods with available codes. 
1) StyleSpeech~\cite{Dongchan2021StyleSpeech} is a multi-speaker voice clone method that synthesizes speech in the style of the target speaker via meta-learning; 
2) FastSpeech2~\cite{ren2020fastspeech} is a popular multi-speaker TTS method for explicitly modeling energy and pitch in speech; 
3) Zero-shot TTS~\cite{YixuanZhou} is a content-dependent fine-grained speaker method for zero-shot speaker adaptation. 
4) V2C-Net~\cite{chen2022v2c} is the first visual voice cloning model for movie dubbing; 
5) HPMDubbing~\cite{cong2023learning} is a hierarchical prosody modeling for movie dubbing, which bridges video representations and speech attributes from three levels: lip, facial expression, and scene. 
6) Face-TTS~\cite{lee2023imaginary} is a novel face-styled speech synthesis within a diffusion model, which leverages face images to provide a robust characteristic of speakers. 
In addition, for a fair comparison, for the pure TTS method, we adopt the setting as~\cite{chen2022v2c}, which takes video embedding as an additional input, before the duration predictor to predict the duration.

\begin{table}[!tbp]
  \centering
  \resizebox{1.0\linewidth}{!}
  {
    \begin{tabular}{lccc}
    \toprule
    \# & Whisper's Version	   & WER on  V2C-Animation (\%) $\downarrow$   \\
    \midrule
    1 & Whisper (base)  &  45.58  \\
    2 & Whisper (large-v1)   & 23.88    \\
    3 & Whisper (large-v2) & 23.85   \\
    4 & Whisper (large-v3)  & \textbf{22.55}   \\
    \bottomrule
    \end{tabular}%
    } 
  \caption{The WER test (the ground truth result) for various versions of Whisper on the V2C benchmark dataset. } 
  \vspace{-10pt}
  \label{tab_whisper}%
\end{table}

\section{Whisper test on V2C-Animation dataset}\label{Whisper}

In Table~\ref{tab_whisper}, the results show that large-v3 achieved the lowest WER, thus we re-selected large-v3 as the measurement tool to get more convincing results on the V2C-Animation dataset.    
Note that Whisper-Large V3 has not been fine-tuned in the V2C-Animation dataset, there is still some gap (WER in GT is 22.55 \%), but it is enough to serve as a fair comparison. 
All results (Dub1.0, 2,0, and 3.0) still reflect that our StyleDubber is the best (see Table~1,2,4) and is more conducive to the clarity of movie dubbing. 
Considering the inference speed and computation memory, the Grid dataset still retains the original 
``Whisper-base'' as the test benchmark. 
The ``Whisper-base'' achieves the 22.41 \% GT WER on the GRID test as similar to the VDTTS~\cite{hassid2022more} result in Table 2 (GRID evaluation).

\begin{table}[!tbp]
  \centering
  \resizebox{1.0\linewidth}{!}
  {
    \begin{tabular}{lccccc}
    \toprule
    Methods & Visual & Sim-O $\uparrow$ & Sim-R $\uparrow$ \\

    \midrule
    Ground Truth & - & 0.79 & n/a   \\
    \midrule
    Fastspeech2~\cite{ren2020fastspeech} & \text{\sffamily X} & 0.10 & 0.19   \\
    StyleSpeech~\cite{Dongchan2021StyleSpeech} & \text{\sffamily X} & 0.14 & 0.23 \\
    Zero-shot TTS~\cite{YixuanZhou} & \text{\sffamily X}  & 0.12 & 0.21 \\
    \midrule
    Fastspeech2*~\cite{ren2020fastspeech} &  \checkmark  &0.10 &0.18 \\ 
    StyleSpeech*~\cite{Dongchan2021StyleSpeech} &  \checkmark   &0.14 &0.23 \\
    Zero-shot TTS*~\cite{YixuanZhou}  &  \checkmark  & 0.13 & 0.22\\
    V2C-Net~\cite{chen2022v2c}  & \checkmark  & 0.08 & 0.15\\
    HPMDubbing~\cite{cong2023learning} &  \checkmark  & 0.11 & 0.19 \\
    Face-TTS~\cite{lee2023imaginary} & \checkmark   & 0.09 & 0.12\\
     \midrule 
    Ours  & \checkmark &  \textbf{0.25} & \textbf{0.34} \\
    \bottomrule
    \end{tabular}%
    } 
  \caption{The V2C-Animation dataset results under the WavLM-TDNN similarity testing.} 
  \vspace{-10pt}
  \label{sotaspksim_V2C}%
\end{table}

\section{WavLM-TDNN Similarity Testing}\label{spksim}

We employ the SOTA speaker verification model, WavLM-TDNN, to evaluate the speaker similarity between the prompt (\ie, the reference audio in the V2C task) and synthesized speech, following VALL-E~\cite{wang2023neural}, VoiceBox~\cite{le2023voicebox}, and NaturalSpeech 3~\cite{ju2024naturalspeech}. 
WavLM-TDNN achieved the top rank at the VoxSRC Challenge 2021 and 2022 leaderboards and it is suitable as the SPK-SIM metric for the  challenging V2C-nimation 
 dataset~\cite{chen2022v2c}. 
The similarity score predicted by WavLM-TDNN is in the range of [-1; 1], where a larger value indicates a higher similarity of input samples. 
Specifically, two metrics need to be calculated: (1) SIM-R represents the similarity with resynthesized audio, which is not comparable across models using different vocoders;
(2) SIM-O is used to measure similarity against the original reference audio. 
Note that the report results based on the GE2E model (Table 1, 2, 4) are compared with the original waveform.

\begin{table}[!tbp]
  \centering
  \resizebox{1.0\linewidth}{!}
  {
    \begin{tabular}{lccccc}
    \toprule
    Methods & Visual & Sim-O $\uparrow$ & Sim-R $\uparrow$ \\

    \midrule
    Ground Truth & - & 0.87 & n/a   \\
    \midrule
    Fastspeech2~\cite{ren2020fastspeech} & \text{\sffamily X} & 0.38 & 0.42   \\
    StyleSpeech~\cite{Dongchan2021StyleSpeech} & \text{\sffamily X} & 0.74 & 0.79 \\
    Zero-shot TTS~\cite{YixuanZhou} & \text{\sffamily X}  & 0.70 & 0.75 \\
    \midrule
    Fastspeech2*~\cite{ren2020fastspeech} &  \checkmark  & 0.48 & 0.52 \\ 
    StyleSpeech*~\cite{Dongchan2021StyleSpeech} &  \checkmark   &0.74 &0.79 \\
    Zero-shot TTS*~\cite{YixuanZhou}  &  \checkmark  & 0.69 & 0.74\\
    V2C-Net~\cite{chen2022v2c}  & \checkmark  & 0.43 & 0.55\\
    HPMDubbing~\cite{cong2023learning} &  \checkmark  & 0.46 & 0.56 \\
    Face-TTS~\cite{lee2023imaginary} & \checkmark   & 0.42 & 0.51 \\
     \midrule 
    Ours  & \checkmark &  \textbf{0.75} & \textbf{0.80} \\
    \bottomrule
    \end{tabular}%
    } 
  \caption{The GRID dataset results under the WavLM-TDNN similarity testing.} 
  \vspace{-10pt}
  \label{sotaspksim_GRID}%
\end{table}

As shown in Table 7,8, the results on two datasets show that even if the similarity measurement method is replaced, our StyleDubber still achieves the best performance in Sim-O and Sim-R. 
The result proves the effectiveness of StyleDubber, which proposes multi-scale style learning at phoneme and utterance levels and captures precise pronunciation in acoustic details and visual emotion for dubbing. 
Besides, we have several findings: (1) Compared with the GT result on the GRID dataset (0.87), the Sim-O result of V2C-Animation is lower (0.79), which may be due to the influence of noise and background music. 
In contrast, the GRID dataset is recorded in a studio environment. 
(2) 
Changing the measurement metric has relatively little impact on the GRID dataset. 
V2C has a very obvious decline in WavLM-TDNN similarity, which is not captured by the GE2E model (GE2E score can reach more than 80\%). 
In the future, we will investigate more robust timbre extractors and use the denoising diffusion probabilistic models to further improve the generated wave quality.

\end{document}